\documentclass{article}
\usepackage[preprint]{neurips_2026}  
\workshoptitle{The Agentic Web}
\usepackage[utf8]{inputenc} 
\usepackage[T1]{fontenc}    
\usepackage{hyperref}       
\usepackage{url}            
\usepackage{booktabs}       
\usepackage{amsfonts}       
\usepackage{nicefrac}       
\usepackage{microtype}      
\usepackage{xcolor}         
\usepackage{amsmath,amssymb,graphicx}
\usepackage{subcaption}     
\usepackage{tikz}\usetikzlibrary{shapes.geometric,shapes.symbols,arrows.meta}
\newcommand{\synapse}{\textsc{Synapse}}
\newif\iffinal \finalfalse
\makeatletter
\newcommand{\defnum}[2]{\expandafter\gdef\csname num@#1\endcsname{#2}}
\newcommand{\num}[1]{%
  \ifcsname num@#1\endcsname\csname num@#1\endcsname
  \else\iffinal\errmessage{Unfilled number: #1}\else\colorbox{red!20}{\texttt{\small\detokenize{#1}}}\fi\fi}
\makeatother
\IfFileExists{tables/numbers.tex}{
\defnum{leak_removed}{1,916}
\defnum{eval_queries_with_exact_match_in_pool}{729}
\defnum{eval_queries_with_near_dup_in_pool}{319}
\defnum{eval_queries_removed}{0}
\defnum{tau_harness_tree_docs}{07ec25f0}
\defnum{tau_harness_tree_matched}{bb344928}
\defnum{tau_harness_tree_current}{18a0d08a}
\defnum{tau_centralized_step}{0.672}
\defnum{tau_centralized_task}{0.358}
\defnum{tau_compendium_gain_pts}{6.7}
\defnum{tau_docs_only_step}{0.679}
\defnum{tau_docs_only_task}{0.364}
\defnum{tau_plain_step}{0.596}
\defnum{tau_plain_task}{0.297}
\defnum{tau_steps_per_seed}{264}
\defnum{tau_synapse_step}{0.663}
\defnum{tau_synapse_task}{0.370}
\defnum{registry_tools}{3{,}180}
\defnum{junk_removed}{192}
\defnum{items_per_client}{5{,}000}
\defnum{c22_seeds}{3}
\defnum{c22_n_queries}{258}
\defnum{conflict_gain_pts}{7.4}
\defnum{conflict_rate_max}{60\%}
\defnum{typed_flat_gap_clean_pts}{8.5}
\defnum{c22_typed_clean}{0.583}
\defnum{c22_flat_clean}{0.497}
\defnum{c22_typed_max}{0.575}
\defnum{c22_flat_max}{0.501}
\defnum{c22_flat_recall}{0.81--0.83}
\defnum{c22_typed_recall}{0.76--0.77}
\defnum{c22_targeted_typed}{0.629}
\defnum{c22_targeted_flat}{0.486}
\defnum{c22_targeted_n}{253}
\defnum{bytes_per_round}{20\,MB}
\defnum{latency_retrieval_ms}{1.1}
\defnum{latency_rerank_ms}{36.6}
\defnum{latency_total_ms}{37.8}
\defnum{gap_pts}{1.1}
\defnum{pooled_gain_min_pts}{12}
\defnum{pooled_gain_max_pts}{27}
\defnum{exp_gain_g1_pts}{4--8}
\defnum{sy_all}{0.555}
\defnum{ce_all}{0.550}
\defnum{do_all}{0.528}
\defnum{lo_all}{0.394}
\defnum{qc_clean_acc}{0.817}
\defnum{qc_over_router_pts}{26}
\defnum{qc_fit_n}{25,000}
\defnum{tu_n}{1--4}
\defnum{tu_g1tool_cov}{95/95}
\defnum{tu_g1cat_cov}{77/77}
\defnum{cluster_threshold}{0.85}
\defnum{retrieval_k}{5}
\defnum{oracle_synapse}{0.762}
\defnum{oracle_central}{0.770}
\defnum{oracle_docs}{0.658}
\defnum{oracle_gap_to_classifier_pts}{5}
\defnum{oracle_exp_gain_pts}{10}
\defnum{oracle_retrieval_share_pts}{21}
\defnum{oracle_insert_frac_single}{0.30--0.36}
\defnum{rs_typed_str}{0.583}
\defnum{rs_typed_flat}{0.460}
\defnum{rs_flat_str}{0.531}
\defnum{rs_flat_flat}{0.497}
\defnum{rs_seeds}{3}
\defnum{rs60_typed_flat}{0.416}
\defnum{multitool_rule}{a prediction is correct if it matches any gold tool of the query, and recall@5 is satisfied if any gold tool is among the five candidates}
\defnum{oracle_exp_gain_min_pts}{7}
\defnum{oracle_exp_gain_max_pts}{12}
}{}
\title{Typed Federated Artifacts for the Agentic Web:\\Sharing Tool-Routing Knowledge Across Frozen, Heterogeneous LLM Agents}
\author{%
  Abhijit Chakraborty \\
  Arizona State University \\
  {\small \texttt{achakr40@asu.edu}}
  \And
  Ni Trieu \\
  Arizona State University \\
  \And
  Vivek Gupta \\
  Arizona State University \\
  {\small \texttt{vgupt140@asu.edu}}
}
\begin{document}
\maketitle

\begin{abstract}
Agents on an open, networked web will run frozen models from different vendors, keep their
histories private, and still have something to teach one another: which tool to call, and when not
to. Weights and adapters cannot carry that knowledge across architectures, and flat text (prompts,
example pools) leaves the protocol unable to say which part is a statistic to noise, which is a rule
that must survive a merge, and which is documentation. We propose exchanging \emph{typed federated
artifacts}, schema-validated objects whose fields make per-field privacy (specified here, not
measured), per-field conflict resolution, and cross-model transfer well-defined, and instantiate
them as \synapse{}\footnote{https://anonymous.4open.science/r/FederatedRAG-EE50/README.md}, a shared compendium of tool-routing knowledge. On StableToolBench
(\num{registry_tools} tools, after removing \num{junk_removed} junk entries and
\num{leak_removed} training items that duplicate or nearly duplicate test queries), a federated
compendium routes within \num{gap_pts} points of a centralized one at \num{bytes_per_round} of JSON
per client per round. The same experience merged and shown to the router as typed fields rather
than one flat string is worth \num{typed_flat_gap_clean_pts} points on clean data and
\num{conflict_gain_pts} under \num{conflict_rate_max} injected contradiction; crossing merge and
rendering shows the halves are inseparable (the typed merge shown flat is the worst arm), while
three conflict policies are indistinguishable, so the conflict log that motivated this work is not
what carries the result. On $\tau$-bench retail every compendium arm lifts a GPT-4o agent's
per-step tool-call accuracy by at least \num{tau_compendium_gain_pts} points, a gain we trace to
format rather than federated experience. Two cautionary findings close the paper: on a
topic-labeled math proxy and on StableToolBench itself, a TF-IDF classifier over the same labeled
experience beats every LLM routing arm (by 48 and \num{qc_over_router_pts} points, most of it
retrieval recall), because the benchmark's pool holds labeled queries for every supposedly unseen
tool and held every test query verbatim before our filter; it cannot measure routing to tools
without labels, which is the case routing exists for.
\end{abstract}

\section{Introduction}\label{sec:intro}
The agentic web does not yet have a unit of exchange. When one organization's agent learns that a
weather API is right for forecasts but wrong for historical data, there is no agreed way to hand
that knowledge to an agent run by someone else on a different model. Sharing model updates is impossible when the models are frozen and differ across organizations;
sharing raw interaction traces leaks private data; sharing flat text, such as prompts or example
pools, leaves the protocol blind: it cannot tell which part of
the text is a numeric statistic to be clipped and noised for privacy, which part is a rule
(``never use this tool for $x$'') that must survive being merged with other contributions, and
which part is canonical documentation.

We argue that the exchanged unit should carry a \emph{type signature at the federation boundary}.
A typed federated artifact is an object $C$ with a schema $\mathcal{S}$ that gives every field a
role, a type, and a validation rule. Three operations that
are ill-defined over flat text become well-defined: per-field differential privacy, because
sensitivity is declared rather than estimated; conflict resolution, because disagreement is detected
and kept per field rather than settled by a vote over whole strings; and transfer across models,
because the artifact is read at inference time rather than baked into parameters.

\synapse{} instantiates this for tool routing, choosing which tool an agent should call. Each
client builds a compendium $C=(M,U,P,T,A)$ from its own experience: tool metadata with usage
statistics ($M$), scenarios in which a tool worked ($U$), precautions about when \emph{not} to use
it ($P$), call templates ($T$), and a structured annex ($A$). Edge aggregators merge contributions
with a typed operator that keeps the per-field majority \emph{and} the dissent; the dissent becomes
a typed conflict log that the router reads as ``do not use when'' conditions. Any frozen LLM can
then retrieve from the global compendium and rerank candidates (Fig.~\ref{fig:workflow}, App.~\ref{app:workflow}).

\paragraph{Contributions.} (i) The typed-artifact abstraction and merge operator, in which
per-field privacy and per-field conflict handling are dispatchable operations rather than heuristics
(\S\ref{sec:method}). (ii) A controlled experiment separating \emph{typing} from
\emph{structure}, and merge from rendering, showing where the typed/untyped gap comes from
(\S\ref{sec:conflict}). (iii) Benchmark results on real tool catalogs at \num{bytes_per_round} per
client per round (\S\ref{sec:results}). (iv) Two negative findings to know before building on the same corpora: a supervised
classifier beats LLM routing on a topic proxy and on StableToolBench itself, whose instruction pool
contains its own test set (\S\ref{sec:lessons}).

\section{Typed compendium and merge}\label{sec:method}
A compendium is valid only if every scenario, precaution and template refers to a registered tool
and every numeric field lies in its declared range; invalid contributions are rejected before merge. Each round, an edge aggregator clusters scenarios
per tool (cosine $\ge$ \num{cluster_threshold}), computes a per-field majority within each cluster over
$\{\texttt{scenario},\texttt{precaution},\texttt{annex}\}$, and attaches whatever disagreed as a
typed \texttt{conflict\_log}, rendered to the router in the same round. Two ablations isolate the mechanism: \textsc{RoundDelayed} holds dissent
back until the next round, and \textsc{Majority} discards it. The structured-but-untyped control
serializes the same fields into one flat JSON string and merges by whole-string majority; there a
conflict log is \emph{not expressible}, since no precaution slot exists to attach dissent to. The schema supports clipping and Laplace noise on the numeric fields of $M$
($(\varepsilon,0)$-differential privacy on the numeric path) and masking of text fields; both are
specified in App.~\ref{app:threat}, and \emph{noise was off in every run reported here}, so no number below is a
privacy--utility measurement. At inference, the router retrieves the top-$k$=\num{retrieval_k}
\emph{distinct tools} over canonical descriptions and scenarios (Jina v2 embeddings, pinned
commit), and a frozen LLM scores the candidates, presented description-first with their ``when to
use'' and ``do not use when'' fields.

\section{Does typing matter, and why? Isolating type from structure}\label{sec:conflict}
\begin{figure}[t]
\begin{subfigure}[c]{.37\linewidth}\centering
\includegraphics[width=\linewidth]{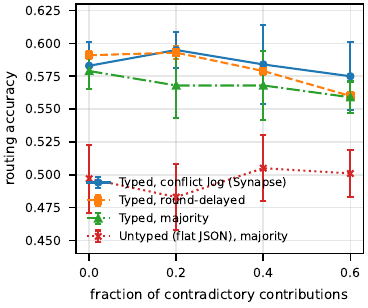}
\caption{}\label{fig:conflict_curve}
\end{subfigure}\hfill
\begin{subfigure}[c]{.61\linewidth}\centering\footnotesize
\setlength{\tabcolsep}{2.5pt}
\begin{tabular}{lcccc}
\toprule
Arm & 0\% & 20\% & 40\% & 60\% \\
\midrule
\synapse{} (typed, log) & $.583{\scriptstyle\pm.018}$ & $.595{\scriptstyle\pm.014}$ & $.584{\scriptstyle\pm.030}$ & $.575{\scriptstyle\pm.026}$ \\
Typed, delayed & $.591{\scriptstyle\pm.003}$ & $.593{\scriptstyle\pm.005}$ & $.579{\scriptstyle\pm.003}$ & $.560{\scriptstyle\pm.003}$ \\
Typed, majority & $.579{\scriptstyle\pm.014}$ & $.568{\scriptstyle\pm.025}$ & $.568{\scriptstyle\pm.026}$ & $.559{\scriptstyle\pm.012}$ \\
Untyped, majority & $.497{\scriptstyle\pm.026}$ & $.483{\scriptstyle\pm.025}$ & $.505{\scriptstyle\pm.025}$ & $.501{\scriptstyle\pm.018}$ \\
Untyped, log & \multicolumn{4}{c}{\emph{not expressible}} \\
\midrule
recall@5 typed / untyped & .764 / .811 & .769 / .808 & .758 / .828 & .769 / .822 \\
seeds (delayed) & 3 (2) & 2 (2) & 3 (2) & 3 (2) \\
\bottomrule
\end{tabular}
\caption{}\label{tab:conflict_2x2}
\end{subfigure}
\caption{Typed vs.\ untyped (flat JSON) under injected contradiction on StableToolBench
G1+G2-instruction ($n$=\num{c22_n_queries}; mean$\pm$SD over seeds 42/123/456, round-delayed and the
20\% column from two seeds; identical injected set across arms per (rate, seed), verified by hash).
(a) Routing accuracy vs.\ fraction of contradictory contributions. (b) The same data with recall@5
and seed counts; an untyped conflict log has no precaution field to attach dissent to. The
typed/untyped gap is already present at 0\%.}\label{fig:conflict}
\end{figure}

We inject contradictions before merge as scenario/precaution pairs about real tools (``use $t$ for
$x$'' versus ``do not'') at rates $\{0,0.2,0.4,0.6\}$ of client
contributions, identically across arms for each (rate, seed), with a candidate pool of 20 (5 in
\S\ref{sec:results}); at 40\% it touches 2,709 of
\num{registry_tools} tools and the gold tool of \num{c22_targeted_n}/\num{c22_n_queries} test queries.
Fig.~\ref{fig:conflict} shows a result we did not predict: the gap is already there at 0\%. The same fields, serialized into one string, cost the reranker
\num{typed_flat_gap_clean_pts} points before a single contradiction is injected
(\num{c22_typed_clean} against \num{c22_flat_clean}, \num{c22_seeds} seeds). Contradiction then
costs the typed arms 1 to 3 points by 60\% and the untyped arm nothing, leaving every typed arm
at 60\% above the untyped arm at 0\%. The untyped arm's
recall@5 (is the correct tool among the five candidates?) is higher throughout, and on the
\num{c22_targeted_n} seed-456 queries whose gold tool was contradicted the typed arms still lead
(\num{c22_targeted_typed} vs.\ \num{c22_targeted_flat}): the typed merge costs retrieval, yet the
gap is in the reranker. Is it the typed \emph{merge} or the typed \emph{rendering}? We crossed them (App.~\ref{app:setup},
Tab.~\ref{tab:renderswap}): the typed merge shown to the model as a flat blob scores
\num{rs_typed_flat}, \emph{below} the flat merge shown flat (\num{rs_flat_flat}); the flat merge
shown with field labels recovers only part of the gap (\num{rs_flat_str}); the typed merge shown
typed scores \num{rs_typed_str}; at 60\% contradiction the flat-shown typed merge falls furthest
(\num{rs60_typed_flat}), since a conflict log read as a blob is noise and read as a field is a
``do not use when''. Neither half works alone: the type has to survive from merge to inference,
which is what a signature at the federation boundary means (shown for one 8B reranker). It matters more than the
conflict-resolution policy: keeping dissent as a conflict log, delaying it a round, or discarding
it gives results within noise of one another at every rate (Fig.~\ref{tab:conflict_2x2}). We do not claim the conflict log
buys accuracy here; it remains the operation only a typed unit can \emph{express}, and measuring
its value needs a benchmark whose gold decisions a contradiction can change, which
StableToolBench's description-driven queries are not.

\section{Benchmarks and transfer}\label{sec:results}
\begin{table}[t]
\centering\small
\caption{{\small StableToolBench} routing accuracy (alias-collapsed), mean over 3 seeds; $K$=5 clients,
\num{items_per_client} items per client from the leak-filtered pool; SDs and recall@5 in App.~\ref{app:setup}.
$^\dagger$ unseen relative to ToolLLaMA's training split only: the instruction pool holds labeled queries for every test tool (\S\ref{sec:lessons}). $^\ddagger$ TF-IDF+SVM on the same \num{qc_fit_n} labeled items per seed, no federation; 0 on any tool without labeled queries. Bold: best compendium arm.}\label{tab:benchmarks}
\hspace{2em}
\setlength{\tabcolsep}{3pt}
\begin{tabular}{lcccccc|c}
\toprule
 & \multicolumn{3}{c}{G1 (single tool)} & \multicolumn{2}{c}{G2 (multi-tool)} & G3 & \\
Arm & instr.\ (153) & tool$^\dagger$ (152) & cat.$^\dagger$ (134) & instr.\ (105) & cat.$^\dagger$ (124) & instr.\ (61) & all (729) \\
\midrule
\textsc{Local-only}      & $0.484$ & $0.457$ & $0.464$ & $0.300$ & $0.278$ & $0.252$ & $0.394$ \\
\textsc{Docs-only}       & $0.562$ & $0.546$ & $0.537$ & $\mathbf{0.533}$ & $\mathbf{0.460}$ & $0.508$ & $0.528$ \\
\synapse{} (federated)   & $\mathbf{0.601}$ & $\mathbf{0.594}$ & $\mathbf{0.614}$ & $0.514$ & $0.438$ & $0.519$ & $\mathbf{0.555}$ \\
\textsc{Centralized}     & $0.599$ & $0.583$ & $\mathbf{0.614}$ & $0.495$ & $0.435$ & $\mathbf{0.530}$ & $0.550$ \\
\midrule
\textsc{Query classifier}$^\ddagger$ & $0.837$ & $0.827$ & $0.838$ & $0.816$ & $0.780$ & $0.771$ & $0.817$ \\
\bottomrule
\end{tabular}
\end{table}

\begin{table}[t]\centering\small
\caption{$\tau$-bench retail, 55 held-out tasks, GPT-4o (\texttt{gpt-4o-2024-08-06}) agent and
simulator, mean$\pm$SD over 3 seeds. Every compendium arm is $\ge$\num{tau_compendium_gain_pts} points
above \textsc{Plain}; the three compendium arms are within one SD. \textsc{Plain} is our own agent
loop in the vendored harness and is not comparable to the leaderboard (App.~\ref{app:setup}).}\label{tab:transfer}
\begin{tabular}{lccc}
\toprule
Arm & Tool-call step acc. & Task success & Steps / seed \\
\midrule
\textsc{Plain} (no compendium) & $0.596\pm0.036$ & $0.297\pm0.046$ & 264 \\
\textsc{Docs-only}             & $0.679\pm0.012$ & $0.364\pm0.048$ & 264 \\
\synapse{} (federated)         & $0.663\pm0.025$ & $0.370\pm0.010$ & 264 \\
\textsc{Centralized}           & $0.672\pm0.019$ & $0.358\pm0.038$ & 264 \\
\bottomrule
\end{tabular}
\end{table}

\textbf{Setup.} $K$=5 clients with a category-coherent partition, \num{items_per_client}
experience items per client drawn from the ToolBench training instructions \emph{after} removing
all \num{leak_removed} items that exactly or nearly ($\cos\ge0.95$) duplicate a test query; three
rounds; router Llama-3.1-8B-Instruct. A TF-IDF+SVM \textsc{Query classifier} trained on the same
labeled items is included because it is what topic proxies reward (\S\ref{sec:lessons}). Latency is in App.~\ref{app:latency}.
\textbf{Results (Tab.~\ref{tab:benchmarks}).} Federation costs nothing: \synapse{} is never more than
\num{gap_pts} points below \textsc{Centralized} and is above or level with it on five of six groups. Pooling
is what matters: \textsc{Local-only} trails by \num{pooled_gain_min_pts} to
\num{pooled_gain_max_pts} points, most on multi-tool queries, where one client's experience covers
few of the tools needed. Experience helps where descriptions run out: relative to
\textsc{Docs-only}, client experience adds \num{exp_gain_g1_pts} points on single-tool queries,
including the splits the benchmark labels unseen (but see \S\ref{sec:lessons}), and about 2
points less on multi-tool queries. That deficit is retrieval, not the artifact: with the gold tool
forced into the five candidates (oracle retrieval, App.~\ref{app:setup}), experience beats
\textsc{Docs-only} by \num{oracle_exp_gain_min_pts} to \num{oracle_exp_gain_max_pts} points on every group, multi-tool included
(\num{oracle_synapse} vs.\ \num{oracle_docs}).
\textbf{$\tau$-bench: where the gain comes from.} On $\tau$-bench retail (14 tools, 55 held-out
tasks, GPT-4o agent and simulator; App.~\ref{app:setup}) the primary metric is per-step
tool-call accuracy (\num{tau_steps_per_seed} decisions per seed); task success on 55 tasks cannot
separate the arms. The plain tool-calling agent reaches \num{tau_plain_step}; every compendium arm
is at least \num{tau_compendium_gain_pts} points higher, and \textsc{Docs-only}, \synapse{} and
\textsc{Centralized} lie within one standard deviation of one another. On this catalog the gain
comes from the compendium \emph{format}, not from federated experience, which adds nothing
measurable on 14 well-documented tools. One artifact under two models
on one benchmark remains untested.

\section{Two negative results for federated-agent benchmarking}\label{sec:lessons}
\textbf{Topic proxies reward classifiers, not routers.} An earlier version of this work used
GSM8k with topic labels such as \emph{Algebraic Word Problem Solver}. A TF-IDF+SVM
classifier reaches 0.92 there and keeps 0.82 when the routing artifact is randomly relabeled (it
reads the query, not the compendium), while the LLM router reaches 0.44 under four prompts: such
benchmarks measure text classification, not routing.
\textbf{The benchmark's pool contains its test set, and the benchmark is lexically solvable.} The
ToolBench instruction pool contains \emph{every} StableToolBench test query as an exact normalized
string match (\num{eval_queries_with_exact_match_in_pool}/729 after the junk filter; 765/765
before). Our classifier scored 1.000 on the raw pool. After the filter, with zero exact or near-duplicate overlap, it still
scores \num{qc_clean_acc}, which is \num{qc_over_router_pts} points above every LLM routing arm, on
the ``unseen'' splits as well. About \num{oracle_retrieval_share_pts} of those points are retrieval
recall: under oracle retrieval the best LLM arm reaches \num{oracle_central}, still
\num{oracle_gap_to_classifier_pts} points short of the classifier. Those splits are unseen only relative to ToolLLaMA's training split.
The pool holds other generated queries for every test tool, each written from that tool's
description and sharing its vocabulary, which a bag-of-words model over \num{qc_fit_n} labeled
queries learns for every tool. The benchmark thus rewards supervised query
classification over routing, and it cannot pose the case routing exists for, a tool with no
labeled queries: every gold tool of the ``unseen'' splits (\num{tu_g1tool_cov} in G1-tool,
\num{tu_g1cat_cov} in G1-category) has labeled queries in the filtered pool, and only \num{tu_n} of
729 test queries per seed have a gold tool absent from the clients' \num{qc_fit_n} items, where the
classifier scores 0 by construction. We release the overlap filter; both findings join recent audits
of agentic evaluation~\citep{zhu2025agenticbenchmarks,bhat2026benchmarkingbenchmarks}.

\section{Related work and outlook}\label{sec:related}
Adapter federation \citep{ye2024openfedllm,kuang2024federatedscope} needs a shared architecture;
prompt and exemplar federation \citep{chen2025fedtextgrad,wang2025fedicl} exchanges flat text;
FICAL \citep{wu2024fical} exchanges free-text ``knowledge compendiums'' merged by an LLM summarizer,
the untyped version of our unit; federated RAG \citep{addison2024cfedrag} federates indices without
typed fields. Routing follows ToolLLM \citep{qin2023toolllm}; \citet{askin2026federatetherouter} federate
model-selection routers as parameters. What is new is the type signature at the federation boundary:
the artifact, not the model, is the unit of exchange.

\bibliographystyle{plainnat}
\bibliography{refs}
\clearpage
\appendix
\section{One round of \synapse{}}\label{app:workflow}
\begin{figure}[htbp]
\hspace{-10em}\centering\includegraphics[width=1.2\textwidth]{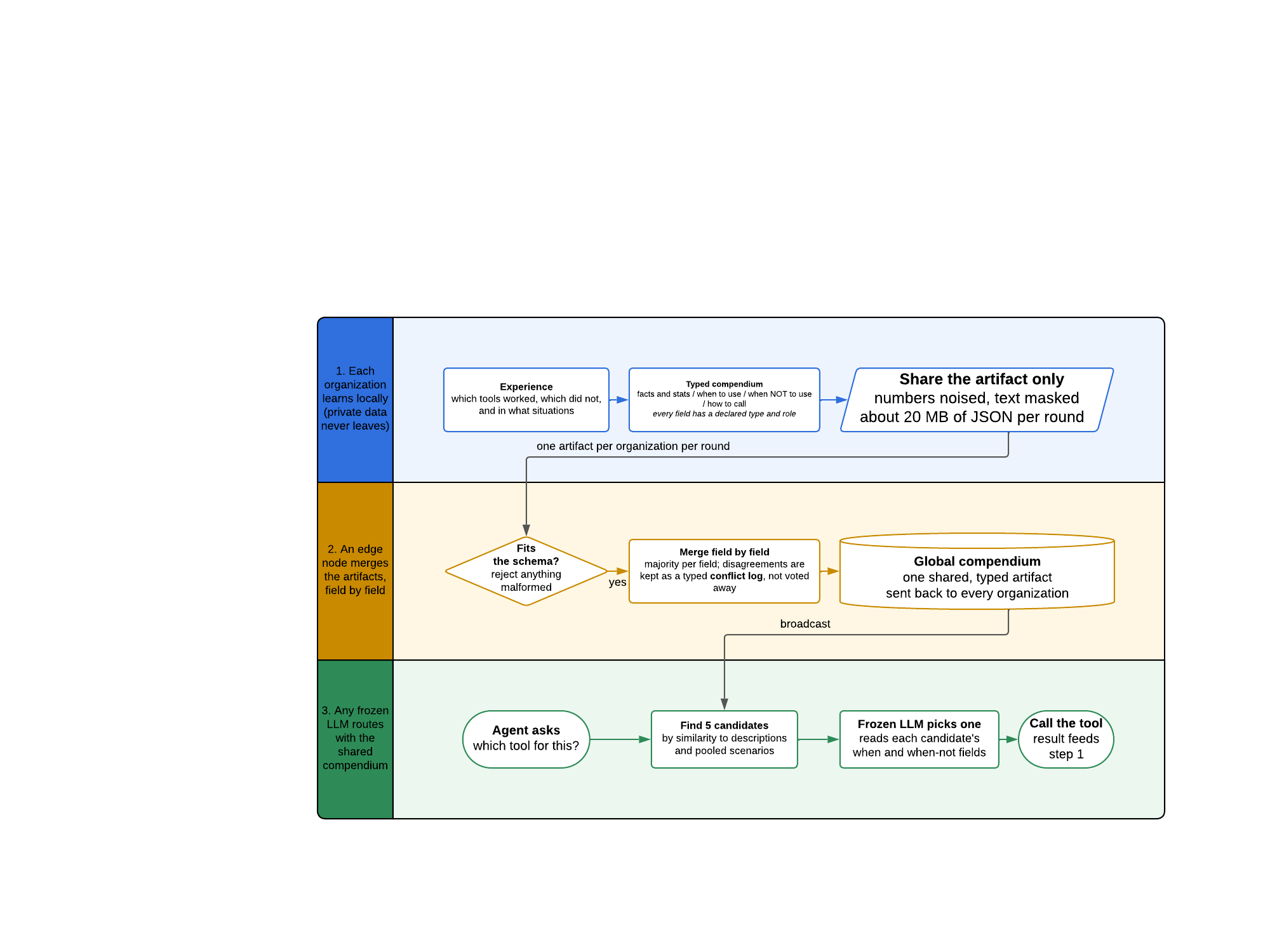}
    \caption{One round of \synapse{}. Each organization turns its private tool-use experience into a typed compendium and shares only that artifact (numeric fields clipped and noised, text masked; about\num{bytes_per_round} of JSON). An edge node validates each artifact against the schema and merges field by field, keeping disagreements as a typed conflict log rather than voting them away, and broadcasts one global compendium. At inference, any frozen LLM retrieves five candidate tools by similarity to descriptions and pooled scenarios and picks one by reading each candidate's ``when'' and ``when not'' fields. The outcome of the call becomes next round's experience.}
    \label{fig:workflow}
\end{figure}
\section{Threat model and privacy mechanisms}\label{app:threat}
\emph{Adversary.} An honest-but-curious edge aggregator and any client that reads the global
compendium; no trusted server. \emph{Numeric path.} Each client's per-tool usage counts and success
rates in $M$ are clipped to a declared range $[0,c]$ and Laplace-noised at the edge with scale
$c/\varepsilon$; over $R$ rounds basic composition gives $(R\varepsilon,0)$-DP for the numeric fields,
and the sensitivity is a property of the schema rather than an estimate, which is the operation
typing makes well-defined. \emph{Text path.} Scenarios and precautions leave clients as short
templated strings after masking of literals (numbers, identifiers, quoted spans); the full paper
measures a prompt-extraction adversary's AUROC against masking strength and against a word-level
metric-LDP alternative. The workshop version makes no formal claim about the text path.
\section{Experimental details and released artifacts}\label{app:setup}
Every number regenerates via \texttt{make tables} from a clean checkout; \texttt{tables/manifest.json}
lists, for each table, the source run directories, their content hashes, and the repository commit each
run recorded. Per-query routing decisions are released as hashed logs.

\begin{table}[h]\centering\scriptsize
\caption{Full StableToolBench results, mean$\pm$SD over seeds 42/123/456 (\textsc{Docs-only} is
deterministic; \textsc{Local-only} is the mean over the five clients, so its SD is over
client-averaged values rather than over one 153-query count). Generated from the three \texttt{combined\_summary.json} files listed in
\texttt{tables/manifest.json}. The query-classifier arm of this run had been fit on the unfiltered pool (0.93--0.98 on every
split) and is excluded; the row in Tab.~\ref{tab:benchmarks} is a separate run fit on the filtered
pool with an asserted zero exact/near-duplicate overlap (\texttt{stabletoolbench\_query\_classifier\_clean\_r1}).}
\label{tab:benchmarks_full}
\setlength{\tabcolsep}{4pt}
\begin{tabular}{lcccccc}
\toprule
& \multicolumn{2}{c}{G1-Instr.\ (153)} & \multicolumn{2}{c}{G1-Tool (152)} & \multicolumn{2}{c}{G1-Cat.\ (134)} \\
Arm & Acc & R@5 & Acc & R@5 & Acc & R@5 \\
\midrule
\synapse{} & $.601\pm.000$ & $.684\pm.008$ & $.594\pm.036$ & $.654\pm.056$ & $.614\pm.024$ & $.684\pm.024$ \\
\textsc{Centralized} & $.599\pm.004$ & $.667\pm.011$ & $.583\pm.034$ & $.634\pm.031$ & $.614\pm.024$ & $.672\pm.020$ \\
\textsc{Local-only} & $.484\pm.002$ & $.556\pm.010$ & $.457\pm.005$ & $.513\pm.002$ & $.464\pm.013$ & $.546\pm.008$ \\
\textsc{Docs-only} & $.562$ & $.706$ & $.546$ & $.691$ & $.537$ & $.709$ \\
\midrule
& \multicolumn{2}{c}{G2-Instr.\ (105)} & \multicolumn{2}{c}{G2-Cat.\ (124)} & \multicolumn{2}{c}{G3-Instr.\ (61)} \\
Arm & Acc & R@5 & Acc & R@5 & Acc & R@5 \\
\midrule
\synapse{} & $.514\pm.019$ & $.619\pm.016$ & $.438\pm.020$ & $.511\pm.017$ & $.519\pm.081$ & $.617\pm.137$ \\
\textsc{Centralized} & $.495\pm.010$ & $.603\pm.036$ & $.435\pm.016$ & $.508\pm.008$ & $.530\pm.077$ & $.612\pm.133$ \\
\textsc{Local-only} & $.300\pm.004$ & $.354\pm.011$ & $.278\pm.019$ & $.322\pm.019$ & $.252\pm.028$ & $.292\pm.018$ \\
\textsc{Docs-only} & $.533$ & $.695$ & $.460$ & $.565$ & $.508$ & $.639$ \\
\bottomrule
\end{tabular}
\end{table}

\paragraph{Oracle retrieval.} To separate retrieval from reranking, we re-evaluate the three
compendium arms with every gold tool of a query forced into the top-5 candidate list, each
replacing the lowest-ranked non-gold candidate (queries needing an insertion: 0.51--0.52 overall,
\num{oracle_insert_frac_single} on single-tool queries). Tab.~\ref{tab:oracle} reports the
result. Experience that was worth 2.7 points under real retrieval is worth
\num{oracle_exp_gain_min_pts}--\num{oracle_exp_gain_max_pts} once the gold tool is present, on every group; the classifier's
remaining margin over the best LLM arm is \num{oracle_gap_to_classifier_pts} points.
\begin{table}[h]\centering\scriptsize
\caption{Routing accuracy with real retrieval (Tab.~\ref{tab:benchmarks}) and with oracle retrieval
(gold tool forced into the five candidates), mean$\pm$SD over seeds 42/123/456; \textsc{Docs-only} is
deterministic. Run \texttt{stabletoolbench\_oracle\_r1}, hashes and commit in
\texttt{tables/manifest.json}.}\label{tab:oracle}
\setlength{\tabcolsep}{2.5pt}
\begin{tabular}{lcc|cccccc}
\toprule
& \multicolumn{2}{c|}{all (729)} & \multicolumn{6}{c}{oracle retrieval, per group} \\
Arm & real retr. & oracle & G1-instr. & G1-tool & G1-cat. & G2-instr. & G2-cat. & G3-instr. \\
\midrule
\synapse{} & $.555$ & $.762\pm.014$ & $.767\pm.023$ & $.750\pm.023$ & $.749\pm.034$ & $.771\pm.019$ & $.750\pm.050$ & $.820\pm.028$ \\
\textsc{Centralized} & $.550$ & $.770\pm.014$ & $.776\pm.016$ & $.752\pm.033$ & $.769\pm.039$ & $.781\pm.019$ & $.753\pm.065$ & $.825\pm.025$ \\
\textsc{Docs-only} & $.528$ & $.658$ & $.667$ & $.632$ & $.657$ & $.657$ & $.637$ & $.754$ \\
\textsc{Query classifier} & $.817$ & -- & \multicolumn{6}{c}{(no retrieval stage)} \\
\bottomrule
\end{tabular}
\end{table}

\paragraph{Merge versus rendering.} Each cell of Fig.~\ref{tab:conflict_2x2} is re-evaluated
with the merge and the rendering crossed: the typed-merged compendium rendered to the reranker as
one flat JSON string, and the flat-merged string decoded back into fields and rendered
description-first. Merged compendiums are regenerated from the same seeds; every regenerated
baseline cell reproduces Fig.~\ref{tab:conflict_2x2} to three decimals. Tab.~\ref{tab:renderswap}
gives the result at 0\% and 60\% contradiction.
\begin{table}[h]\centering\scriptsize
\caption{Merge and rendering crossed (StableToolBench G1+G2-instruction, $n$=\num{c22_n_queries},
mean$\pm$SD over seeds 42/123/456; run \texttt{stabletoolbench\_2x2\_renderswap\_r1}, commit and
per-cell hashes in \texttt{tables/manifest.json}). The two baseline rows reproduce
Tab.~\ref{tab:conflict_2x2} to three decimals. The decoded row re-parses the flat merge's stored
string into fields; its re-serialization is not byte-identical to the stored string, so it is an
approximation of the flat merge's content. The typed merge helps only when the reranker is shown its
fields; shown flat, it is worse than the flat merge, and worst under contradiction.}
\label{tab:renderswap}
\setlength{\tabcolsep}{5pt}
\begin{tabular}{llcc|cc}
\toprule
& & \multicolumn{2}{c|}{0\% contradiction} & \multicolumn{2}{c}{60\% contradiction} \\
Merge & Rendered to reranker as & Acc & R@5 & Acc & R@5 \\
\midrule
typed (conflict log) & typed fields (\synapse) & $.583\pm.018$ & $.764$ & $.575\pm.026$ & $.769$ \\
typed (conflict log) & one flat string & $.460\pm.006$ & $.738$ & $.416\pm.025$ & $.749$ \\
flat (majority)      & typed fields (decoded) & $.531\pm.014$ & $.820$ & $.539\pm.014$ & $.827$ \\
flat (majority)      & one flat string & $.497\pm.026$ & $.811$ & $.501\pm.018$ & $.822$ \\
\bottomrule
\end{tabular}
\end{table}

\paragraph{Routing constants.} Retrieval is \texttt{distinct\_tool\_topk} over Jina-v2
embeddings with $k$=\num{retrieval_k} candidates reranked by the V3 description-first prompt; the
candidate pool before distinct-tool filtering is 20 in the typing-isolation runs (\S\ref{sec:conflict})
and 5 in the benchmark runs (\S\ref{sec:results}). Edge clustering uses cosine $\ge$
\num{cluster_threshold}; the leak filter removes exact matches and $\cos\ge0.95$ near-duplicates and
reports $\cos\ge0.90$. Accuracy is alias-collapsed top-1 tool match; on multi-tool (G2, G3) queries \num{multitool_rule}. No differential-privacy noise was
applied in any reported run.
\paragraph{Contradiction injection.} Each injected contradiction is a scenario \emph{and} a
precaution attributed to different clients, so injection also adds positive scenarios; this is why
recall@5 rises slightly with rate in \S\ref{sec:conflict}. Precaution-only injection is the cleaner
design and is left for the full study.
\paragraph{$\tau$-bench harness provenance.} We vendor \texttt{sierra-research/tau-bench} into the
repository and record the git tree hash of the vendored directory in every run's sidecar. The
\textsc{Docs-only} rows in Tab.~\ref{tab:transfer} were produced on tree \texttt{\num{tau_harness_tree_docs}}
and the \textsc{Plain}/\synapse/\textsc{Centralized} rows on tree \texttt{\num{tau_harness_tree_matched}}.
The diff between the two trees is confined to \texttt{tau\_bench/local\_completion.py} (the local-model
completion path); every run in the table records \texttt{provider: openai} for both agent and
simulator, under which that file is not executed, and \texttt{make tables} admits the mixed trees only
under exactly that condition, with the diff and provider fields recorded in the manifest. A direct
replication of \textsc{Docs-only} on the current tree (\texttt{\num{tau_harness_tree_current}}) is
reported if it completes before the camera-ready. \textsc{Plain} is our own tool-calling agent loop
inside the vendored harness evaluated at pass$^1$ on a 55-task holdout (split hash in the manifest);
its task-success rate is therefore not comparable to the $\tau$-bench leaderboard's pass$^1$ on all 115
retail tasks, and we make no such comparison. All four arms share harness, tasks, simulator, seeds and
grader, and are comparable to one another. Per-step tool-call accuracy is scored by longest-common-subsequence
matching of predicted to gold tool calls within each task.
\section{Scale and latency measurements}\label{app:latency}
At \num{registry_tools} registered tools the retrieval index holds canonical descriptions plus
client scenarios (28,433 entries after merge on seed 456); dense retrieval over it costs
\num{latency_retrieval_ms}\,ms per query and the 8B logit-scoring rerank of five distinct candidates
\num{latency_rerank_ms}\,ms, for \num{latency_total_ms}\,ms end-to-end (means over 258 queries, one
A100-class card, batch size 1). The edge merge, not inference, dominates wall-clock: clustering
28k scenarios takes roughly an hour per round on CPU in our unoptimized implementation, which is
the cost a deployment would pay once per aggregation round rather than per query.
\end{document}